\documentclass[review]{elsarticle}
\usepackage{prletters}
\usepackage{framed,multirow}

\usepackage{amssymb}
\usepackage{latexsym}

\usepackage{url}	
\usepackage{xcolor}
\definecolor{newcolor}{rgb}{.8,.349,.1}

\usepackage{times}	
\usepackage{amssymb}
\usepackage{graphicx}
\usepackage{multirow}
\usepackage{pdfpages}
\usepackage{float}
\usepackage{subcaption}
\usepackage{caption}

\usepackage[utf8]{inputenc}

\usepackage{blindtext}
\usepackage{color,soul}

\usepackage{verbatim}

\usepackage[acronym]{glossaries}
\usepackage{algorithm}
\usepackage{algorithmicx}
\usepackage{pdfpages}
\usepackage{time}
\usepackage{appendix}
\usepackage{amsmath}
\usepackage{amssymb}
\usepackage{sidecap}

\DeclareMathOperator*{\argmin}{arg\,min}
\DeclareMathOperator*{\argmax}{arg\,max}

\newacronym{pos}{POS}{Part-Of-Speech}
\newacronym{nlp}{PLN}{Processamento da Língua Natural}
\newacronym{svm}{SVM}{Support Vector Machine}
\newacronym{pca}{PCA}{Principal Component Analysis}
\newacronym{mfcc}{MFCC}{Mel-Frequency Cepstrum Coeficient}
\newacronym{bsr}{RSB}{Regra da Soma de Bayes}
\newacronym{me}{ME}{Máxima Entropia}
\newacronym{ner}{NER}{Named-Entity Recognition}

\newacronym{LSA}{LSA}{Latent Semantic Analysis}
\newacronym{TF-IDF}{TF-IDF}{Term Frequency - Inverse Document Frequency}
\newacronym{SVD}{SVD}{Singular Value Decomposition}

\newacronym{MMR}{MMR}{Maximal Marginal Relevance}
\newacronym{GRASSHOPPER}{GRASSHOPPER}{Graph Random-walk with Absorbing StateS
that HOPs among PEaks for Ranking}
\newacronym{ROUGE}{ROUGE}{Recall-Oriented Understudy for Gisting Evaluation}
\newacronym{KP-Centrality}{KP-Centrality}{Key Phrase-based Centrality}

\newacronym{gmm}{GMM}{Gaussian mixture model}
\newacronym{gmms}{GMMs}{Gaussian mixture models}
\newacronym{EM}{EM}{Expectation Maximization}
\newacronym{kde}{KDE}{Kernel density estimation}
\newacronym{okde}{oKDE}{online kernel density estimation}
\newacronym{pdf}{pdf}{probability density function}
\newacronym{oisvm}{OISVM}{online independent support vector machine}

\def\transpose{{\sf T}}

\journal{Pattern Recognition Letters}

\begin{document}

\begin{frontmatter}

\title{Fast and Extensible Online Multivariate Kernel Density Estimation}

\author[1,3]{Jaime \snm{Ferreira}}
\author[1,3]{David \snm{Martins de Matos}\corref{cor1}}
\ead{david.matos@inesc-id.pt}
\cortext[cor1]{Corresponding author}
\author[1,2]{Ricardo \snm{Ribeiro}}

\address[1]{L2F - INESC ID Lisboa, Rua Alves Redol, 9, 1000-029 Lisboa, Portugal}
\address[2]{Instituto Universitário de Lisboa (ISCTE-IUL), Av. das Forças Armadas, 1649-026 Lisboa, Portugal}
\address[3]{Instituto Superior Técnico, Universidade de Lisboa, Av. Rovisco Pais, 1049-001 Lisboa, Portugal}

\received{00 May 2015}
\finalform{00 May 2015}
\accepted{00 May 2015}
\availableonline{00 May 2015}
\communicated{XXX}

\begin{abstract}

In this paper we present xokde++, a state-of-the-art online kernel density estimation approach that maintains Gaussian mixture models input data streams. The approach follows state-of-the-art work on online density estimation, but was redesigned with computational efficiency, numerical robustness, and extensibility in mind. Our approach produces comparable or better results than the current state-of-the-art, while achieving significant computational performance gains and improved numerical stability. The use of diagonal covariance Gaussian kernels, which further improve performance and stability, at a small loss of modelling quality, is also explored. Our approach is up to 40 times faster, while requiring 90\% less memory than the closest state-of-the-art counterpart.

\end{abstract}


\end{frontmatter}

%\linenumbers

%%%%%%%%%%%%%%%%%%%%%%%%%%%%%%%%%%%%%%%%%%%%%%%%%%%%%%%%%%%%%%%%%%%%%%%%%%%%%%%%
%%%%%%%%%%%%%%%%%%%%%%%%%%%%%%%%%%%%%%%%%%%%%%%%%%%%%%%%%%%%%%%%%%%%%%%%%%%%%%%%

\section{Introduction}

Online learning is needed when data is not known a priori, its distribution evolves, or when the data model must be updated without interrupting operation: the model must evolve as new data appears, allowing older data to become progressively less important. Our goal is to produce a fast, robust way of handling long-running modelling tasks. Since capturing, processing, and saving all past observed samples is computationally unfeasible, similar samples can be grouped and represented by more compact entities, such as \gls{gmms}.

\gls{kde} methods are nonparametric approaches to build \gls{gmms} that do not require prior definition of the number of components. 
In order to manage complexity, some approaches compress the model to a predefined number of components~\cite{goldberger2004hierarchical}, or optimize some data-driven choice~\cite{chen2010probability}. A different approach~\cite{ozertem2008mean} is to view model compression as a clustering problem. The main difficulty of adapting \gls{kde} methods to online operation lies in maintaining sufficient information to generalize to unobserved data and to adjust model complexity without access to all past samples. An online approach based on mean-shift mode finding adds each new sample to the model as a Gaussian component. However, it is sensitive to non-Gaussian areas due to skewed or heavy tailed data~\cite{han2008sequential}. A two-level approach~\cite{declercq2008online}, based on the idea that each component of the (non-overfitting) mixture is in turn represented by an underlying mixture that represents data very precisely (possibly overfitting), allows the model to be refined without sacrificing accuracy, as a merge in the upper level can later be accurately split. It is also able handle the problem of non-Gaussian data by using a uniform distribution in those regions.

The \gls{okde} approach~\cite{kristan2014online,kristan2011multivariate} builds a two-level model of the target distribution. It maintains a sample distribution that is used to compute the corresponding \gls{kde}. It supports single-sample updates and has a forgetting factor. This allows model adaptation when the target distribution evolves. Compressions approximate component clusters using single Gaussians. If the target distribution changes sufficiently, compressions can be reversed, using the underlying model kept by each component.

Our approach follows \gls{okde}, obtaining comparable or better results. It allows for flexibility, abstraction, efficient memory use, computational efficiency, and numerical robustness. Furthermore, our C++ implementation, based on state-of-the-art libraries~\cite{eigenweb}, allows easy exploration of modelling alternatives.

The article is organized as follows: Section~\ref{sec:okde} describes \gls{okde}. Section~\ref{sec:numstab} describes our adaptations for numerical stability. Section~\ref{sec:diagcovmat} describes our approach to high dimensionality. Section~\ref{sec:lazy} presents specific computational strategies to improve performance. Section~\ref{sec:eval_setup} describes the evaluation setup and Section~\ref{sec:results} presents our results. Section~\ref{sec:conclusions} concludes the article.

%%%%%%%%%%%%%%%%%%%%%%%%%%%%%%%%%%%%%%%%%%%%%%%%%%%%%%%%%%%%%%%%%%%%%%%%%%%%%%%%
%%%%%%%%%%%%%%%%%%%%%%%%%%%%%%%%%%%%%%%%%%%%%%%%%%%%%%%%%%%%%%%%%%%%%%%%%%%%%%%%

\section{Online Kernel Density Estimation}
\label{sec:okde}

\gls{okde} was not developed with high dimensionality in mind. Although it allows for any number of dimensions, computational complexity scales quadratically with the number of dimensions, making the approach computationally prohibitive. Furthermore, problems due to overflows, underflows, and precision issues also arise. These issues worsen as the number of dimensions grows, limiting the usefulness of the approach. Before describing our approach, we present \gls{okde} in detail.

\subsection{Model Definition}

Sample models are defined as $d$-dimensional, $N$-component \gls{gmms} (Eq.~\ref{eq:gmm}): {\textbf x} are $d$-dimensional observations; $w_i$ are mixture weights, such that $\Sigma_{i=1}^{N}w_i{=}1$; and $\phi_{\Sigma_i}(\text{\textbf x}{-}\mu_i)$ are $d$-variate components (Eq.~\ref{eq:gauss_PDF}), with mean $\mu_i$ and covariance $\Sigma_i$.
\begin{equation}
\label{eq:gmm}
p_{s}(\text{\textbf x})= \sum_{i=1}^{N}w_i \phi_{\Sigma_i}(\text{\textbf x} - {\text{\textbf x}}_{i})
\end{equation}
\begin{equation}
\label{eq:gauss_PDF}
\phi_{\Sigma_i}(\text{\textbf x} - \mu_i) =
\frac{1}{2\pi^{D/2}|\Sigma_i|^{1/2}}\exp\left \{\frac{1}{2}(\text{\textbf x}-\mu_i)^\transpose\Sigma_{i}^{-1}(\text{\textbf x}-\mu_i)\right \}
\end{equation}

To maintain low complexity, $p_{s}(\text{\textbf x})$ is compressed when a threshold is met. To allow recovery from mis-compressions, a model of the observations is also kept for each component:
$\text{\textbf S}_{\text{model}}{=}\{p_{s}(\text{\textbf x}) , \{q_{i}(\text{\textbf x})\}_{i=1:N}\}$, where
$q_{i}(\text{\textbf x})$ is a mixture model (with at most two components) for the $i$-th component of $p_{s}(\text{\textbf x})$.

The kernel density estimate is defined as a convolution of $p_{s}(\text{\textbf x})$ by a kernel with a covariance (bandwidth) {\textbf H}:
\begin{equation}
\hat{p}_{\text{\tiny KDE}}(\text{\textbf x})= \phi_{\text{\textbf H}}(\text{\textbf x}) \ast p_{s}(\text{\textbf x}) =
\sum_{i=1}^{M}w_i \phi_{\text{\textbf H}+\Sigma_i}(\text{\textbf x}
- {\text{\textbf x}}_{i})
\end{equation}

\subsection{Optimal Bandwidth}
\label{sec:opt_band}

KDE approaches determine a bandwidth {\textbf H} that minimizes the distance between $\hat{p}_{\text{\tiny KDE}}(\text{\textbf x})$ and (the unknown) $p(\text{\textbf x})$ generating the data. The bandwidth can be written as $\text{\textbf H}{=}\beta^2F$ ($\beta$ is the scale and $F$ is the structure). Finding {\textbf H} is equivalent to finding the optimal scaling $\beta_{\text{opt}}$ (Eq.~\ref{eq:optscale}), where $R(p,F)\approx\hat{R}(p,F,\text{\textbf G})$ (Eq.~\ref{eq:rpfg_vanilla}) and $\text{\textbf A}_{ij}{=}(\Sigma_{gj}{+}\Sigma_{sj})^{-1}$, $\Delta_{ij}{=}\mu_i{-}\mu_j$, $m_{ij}{=}\Delta_{ij}^\transpose\text{\textbf A}_{ij} \Delta_{ij}$, $\Sigma_{gj}{=}\text{\textbf G}{+}\Sigma_{sj}$, and $\Sigma_{sj}$ is the covariance of the $j$-th component of $p_s(\text{\textbf x})$.
\begin{equation}
\label{eq:optscale}
\beta_{\text{opt}} = [d(4\pi)^{d/2}NR(p,F)]^{-1/(d+4)}
\end{equation}
\begin{eqnarray}
\label{eq:rpfg_vanilla}
\hat{R}(p,F,\text{\textbf G}) = \sum_{i=1}^{N} \sum_{j=1}^{N}w_iw_j
\phi_{\text{\textbf A}_{ij}^{-1}}(\Delta_{ij}) \nonumber \\ \times 
\Big[[2\text{tr}(F^2 \text{\textbf A}_{ij}^2)][1-2m_{ij}] +
\text{tr}^2(F \text{\textbf A}_{ij})[1-m_{ij}]^2 \Big]
\end{eqnarray}

{\textbf G} is the {\em pilot bandwidth} (Eq.~\ref{eq:pilotbandwidth}). $\hat{\Sigma}_{smp}$ is the covariance of a single Gaussian approximating the entire model. The structure of {\textbf H} can be approximated by the structure of the covariance of the observations~\cite{wand1994kernel,duong2003plug}, i.e., $F{=}\hat{\Sigma}_{\text smp}$.

\begin{equation}
\text{\textbf G} = \hat{\Sigma}_{smp} \bigg(\frac{4}{(d+2)N}\bigg)^{2/(d+4)}
\label{eq:pilotbandwidth}
\end{equation}

\subsection{Model Compression}
\label{sec:model_compression}

Compression approximates the original $N$-component $p_{s}(\text{\textbf x})$ by an $M$-component, $M{<}N$, $\hat{p}_{s}(\text{\textbf x})$~(Eq.~\ref{eq:compressed_sp}), such that the compressed \gls{kde} does not change significantly.

\begin{equation}
\label{eq:compressed_sp}
\hat{p}_{s}(\text{\textbf x}) = \sum_{j=1}^{M}\hat{w}_j
\phi_{\hat{\Sigma}_j}(\text{\textbf x} - \hat{\text{\textbf x}}_{j})
\end{equation}

Clustering approaches can be used to compress $p_{s}(\text{\textbf x})$. The aim is to find clusters such that each cluster can be approximated by a single component in $\hat{p}_{s}(\text{\textbf x})$. Let $\Xi(M){=}\{\pi_j\}_{j=1:\text M}$ be a set of assignments clustering $p_{s}(\text{\textbf x})$ into $M$ sub-mixtures. The sub-mixture corresponding to indexes $i{\in}\pi_j$ is defined as 
\begin{equation}
\label{eq:sub_mix_ps}
p_{s}(\text{\textbf x};\pi_j) = \sum_{i\in\pi_j}w_i
\phi_{\Sigma_i}(\text{\textbf x} - {\text{\textbf x}}_{i})
\end{equation}

The parameters of each $\hat{w}_j \phi_{\hat{\Sigma}_j}(\text{\textbf x}{-}\hat{\mu}_{j}) $ are defined by matching the first two moments (mean and covariance) of $\pi_j$
\begin{eqnarray}
\label{eq:moment_match}
\hat{w}_j =  \sum_{i\in\pi_j} w_i , \qquad  
\hat{\mu_j} = \hat{w_j}^{-1}\sum_{i\in\pi_j}w_i\mu_i ,  \\ 
\hat{\Sigma}_j = \bigg(\hat{w}_j^{-1}\sum_{i\in\pi_j}w_i(\Sigma_i + \mu_i\mu_i^\transpose)\bigg) - \hat{\mu_j}\hat{\mu_j}^\transpose\nonumber
\end{eqnarray}

$\Xi(M)$ must be such that $\hat{M}{=}\argmin_M E(\Xi(M))$ and $E(\Xi(\hat{M})){\leq}D_{th}$ (a threshold).
$E(\Xi(M))$ is the largest local clustering error $\hat{E}(p_s(\text{\textbf x}; \pi_j),\text{\textbf H}_{\text{opt}})$, i.e., the error induced under the KDE if $p_s(\text{\textbf x}; \pi_j)$ is approximated by a single Gaussian.
\begin{equation}
\label{eq:local_clust_error_gen}
E(\Xi(M)) = \max_{\pi_j \in \; \Xi(M)} \hat{E}(p_s(\text{\textbf x};
\pi_j),\text{\textbf H}_{\text{opt}})
\end{equation}

\subsection{Local Clustering Error}

Consider $p_1(\text{\textbf x})$, a sub-mixture of $p_s(\text{\textbf x})$, and $p_0(\text{\textbf x})$, its single-Gaussian approximation (Eq.~\ref{eq:moment_match}). 
The local clustering error is the distance (Eq.~\ref{eq:lce}) between the KDEs. It may be quantified by using the Hellinger distance~\cite{pollard2002user} (Eq.~\ref{eq:hellinger_integ}).
\begin{equation}
\label{eq:lce}
\hat{E}(p_1(\text{\textbf x}),\text{\textbf H}_{\text{opt}}) =
D(p_{1\text{KDE}}(\text{\textbf x}),p_{0\text{KDE}}(\text{\textbf x}))
\end{equation}
\begin{equation}
\label{eq:hellinger_integ}
\begin{array}{l c}
D^2(p_{1\text{KDE}}(\text{\textbf x}),p_{0\text{KDE}}(\text{\textbf x})) 
\triangleq \\ \qquad \frac{1}{2}\int\Big(p_{1\text{KDE}}(\text{\textbf x})^{1/2} -
p_{0\text{KDE}}(\text{\textbf x})^{1/2}\Big)^2 d\text{\textbf x}
\end{array}
\end{equation}

\subsection{Distance between mixture models}

The Hellinger distance cannot be calculated analytically for mixture models and is approximated by the unscented transform~\cite{julier1996general} (Eq.~\ref{eq:hell_unscented}), where $\{^{(j)}\mathcal{X}_i,^{(j)}\mathcal{W}_i\}_{j=0:2d+1}$ are weighted sets of sigma points of the $i$-th Gaussian $\phi_{\Sigma_i}(\text{\textbf x}{-}\text{\textbf x}_i)$. 
\begin{equation}
\label{eq:hell_unscented}
D^2(p_1,p_2) \approx \frac{1}{2} \sum_{i=1}^N w_i \sum_{j=0}^{2d+1}
g(^{(j)}\mathcal{X}_i)^{(j)}\mathcal{W}_i
\end{equation}
\begin{equation}
\label{eq:hell_unscented_bits}
\begin{gathered}
^{(0)}\mathcal{X}_i = \text{\textbf x}_i, \quad ^{(0)}\mathcal{W}_i =
\frac{k}{d+k}, \quad k{=}\max([0,m-d])\\
^{(j)}\mathcal{X}_i =  \text{\textbf x}_i + s_j \sqrt{(d+k)\xi_j}, \quad
^{(j)}\mathcal{W}_i = \frac{1}{2(d+k)} \nonumber \\
s_j = 
\begin{cases}  
 1, &  j \leq d \\ -1, & \text{otherwise} 
\end{cases}
\end{gathered}
\end{equation}
Note that $\xi_j$ is the $j$-th column of $\xi{=}\sqrt{\Sigma_i}$ ($\xi$ has $d$ columns). The sigma points are simply $\text{\textbf x}_i $ ($j{=}0$) and $\text{\textbf x}_i{\pm}\xi_j$. Thus, $\xi_j$ must be counted separately for the $s_j{=}1$ and $s_j{=}{-}1$ sets, such that $j{=}[1,d]$ for each set of sigma points. Let $UDU\transpose$ be a singular value decomposition of $\Sigma$, such that $U{=}\{U_1,\dots,U_d\}$ and $D{=}\text{diag}\{\lambda_1,\dots,\lambda_d\}$, then $\xi_k{=}\sqrt{\lambda_k}U_k$. In line with~\cite{julier1996general}, $m$ was set to 3.

\subsection{Hierarchical Compression}

A hierachical approach avoids evaluating all possible assignments $\Xi(M)$~\cite{goldberger2004hierarchical,ihler2005inference}. $p_s(\text{\textbf x})$ is first split into two sub-mixtures using Goldberger's K-means~\cite{goldberger2004hierarchical}. 
To avoid singularities associated with Dirac-delta components of $p_s(\text{\textbf x})$, K-means is applied to $\hat{p}_{\text{\tiny KDE}}(\text{\textbf x})$. Next, each sub-mixture is approximated by a single Gaussian (section~\ref{sec:model_compression}) and the sub-mixture yielding the largest local error is further split into two sub-mixtures. The process continues until $E(\Xi(\hat{M})){\leq}D_{th}$, producing a binary tree where each of the $\hat{M}$ leaves represents the clustering assignments $\Xi(\hat{M}){=}\{\pi_i\}_{j=1:M}$. $\Xi(\hat{M})$ can then be used to approximate the corresponding components in $p_s(\text{\textbf x})$ by a single Gaussian, resulting in the compressed distribution $\hat{p}_s(\text{\textbf x})$.

If two components of $p_s(\text{\textbf x})$ are merged, their detailed models must also be merged: the detailed model $\hat{q}_j(\text{\textbf x})$ of the $j$-th component in $\hat{p}_s(\text{\textbf x})$ is calculated by first defining a normalized extended mixture (Eq.~\ref{eq:extmix}). If this mixture has more than two components, then the two-component $\hat{q}_j(\text{\textbf x})$ is generated by splitting $\hat{q}_{j\text{ext}}(\text{\textbf x})$ into two sub-mixtures using Golberger's K-means and approximating each sub-mixture by a single Gaussian.  
\begin{equation}
\label{eq:extmix}
\hat{q}_{j\text{ext}}(\text{\textbf x}) = \Big( \sum_{i\in\pi_j}w_i\Big)^{-1}
\sum_{i\in\pi_j} w_i \, q_i(\text{\textbf x})
\end{equation}

\subsection{Revitalization}
\label{subsec:revitalization}

Over-compressions can be detected by checking whether the local clustering error $\hat{E}(q_i(\text{\textbf x}),\text{\textbf H}_{\text{opt}})$ of each $p_s(\text{\textbf x})$ component, evaluated against its detailed model $q_i(\text{\textbf x})$, exceeds $D_{\text{th}}$. Those components are removed from $p_s(\text{\textbf x})$ and replaced by the components of their detailed models. Each new component needs a detailed model: new ones are generated based on their covariances. Let $w_i \phi_{\Sigma_i}(\text{\textbf x}{-}\mu_i)$ be one of the new components. If the determinant of $\Sigma_i$ is zero, then the component is a single data-point and its detailed model is just the component itself. Otherwise, the component has been generated through clustering in previous compression steps: its detailed model is initialized by splitting $\phi_{\Sigma_i}(\text{\textbf x}{-}\mu_i)$ along its principal axis~\cite{huber2015nonlinear} into a two-component mixture, whose first two moments match those of the original component. This has two advantages: first, since the component is symmetric around the mean, the splitting process minimizes the induced error; and second, it is moment preserving, i.e., the mean and covariance of the split Gaussian (and, thus, of the entire mixture), remains unchanged.

%%%%%%%%%%%%%%%%%%%%%%%%%%%%%%%%%%%%%%%%%%%%%%%%%%%%%%%%%%%%%%%%%%%%%%%%%%%%%%%%
%%%%%%%%%%%%%%%%%%%%%%%%%%%%%%%%%%%%%%%%%%%%%%%%%%%%%%%%%%%%%%%%%%%%%%%%%%%%%%%%

\section{Improving Numeric Stability}
\label{sec:numstab}

In this section, we detail strategies that allow working with skewed or degenerate sample distributions, as well as avoiding underflows and overflows. 

\subsection{Degenerate Covariance Matrices}
\label{sec:degen_cov_mat}

Some data dimensions may seldom change or sufficient observations with different values may not have occurred, causing covariance values of 0 along those axes. One way to detect and correct singular or near-singular matrices is to compute their eigen decomposition $\Sigma{=}Q\Lambda Q^\transpose$, with $\Lambda_{ii}{=}\lambda_{i}$, and check for eigenvalues smaller than $10^{-9}$. These eigenvalues are then corrected by 1\% of the average of the eigenvalues (Eq.~\ref{eq:lambdaii},~\ref{eq:lambdaiiprime},~\ref{eq:alpha}). The corrected covariance is then given by $\Sigma'{=}Q\Lambda'Q^\transpose$.

\begin{equation}
\label{eq:lambdaii}
\hat{\Lambda}_{ii} = \hat{\lambda}_i = \frac{\lambda_i}{\argmax_i \lambda_i}
\end{equation}
\begin{equation}
\label{eq:lambdaiiprime}
\Lambda'_{ii} = \lambda'_{i} = \begin{cases}
        \lambda_i  \quad \text{if} \: \hat{\lambda}_i \geq 10^{-9}\\
        \alpha  \quad \text{otherwise}
        \end{cases}
\end{equation}
\begin{equation}
\label{eq:alpha}
\alpha = \frac{1}{100} |\{\lambda_j : \hat{\lambda_i} > 10^{-9}\}|^{-1} \sum_{\hat{\lambda_i} > 10^{-9}}\lambda_i, \:  
\end{equation}

\subsection{Determinant Computation}

The Gaussian \gls{pdf} (Eq.~\ref{eq:gauss_PDF}) uses the determinant ($|\Sigma|$) and the inverse ($\Sigma^{-1}$) of the covariance ($\Sigma$). A way to efficiently compute the inverse of a non-singular matrix is to first compute its LU decomposition and use it to compute the inverse. Since we guarantee that $\Sigma$ is positive definite, the determinant can be computed by multiplying the diagonal entries of U. However, when working in high dimensionality, overflows or underflows are bound to occur, even when using double precision floating point. To avoid this problem, we compute the logarithm of the determinant instead: this is simply the sum of the logarithm of each entry in the diagonal of matrix U.

\subsection{Whitening Transformation}

A single $\text{\textbf H}_{\text{opt}}$ means that all components are scaled equally in all dimensions. If the spread of data is much greater in one dimension than in others, extreme differences of spread in the various dimensions may occur. This can be avoided by first whitening the data, by transforming it to have unit covariance, smoothing it using a radially symmetric kernel, and, finally, transforming it back~\cite{fukunaga1990introduction}. Whitening allows $\text{\textbf H}_{\text{opt}}$ to better fit the global distribution of the data, leading to better models~\cite{silverman1986density}.

In our approach, both the computation of $\text{\textbf H}_{\text{opt}}$ and model compression are performed on whitened data. Whitening starts by approximating the whole model by a single Gaussian. The parameters needed to transform the model covariance into the identity matrix are then computed and individually applied to all components in the mixture. The inverse operation can be used to recover the model.

%Whitening is defined as $Y{=}Q^\transpose\text{\bf{X}}$, where $Q$ are the eigenvectors of $\Sigma$, such that $Q^\transpose\Sigma Q{=}I$. If $\Sigma$ is positive definite, $Q$ exists and is non-singular. The whitening transform is thus reversible.
%The whitening transformation of a data vector $\text{\textbf x}$ with covariance $\Sigma$, is computed in two steps. First, we de-correlate the data, making it have diagonal covariance: $\bf{y}{=}\Phi^\transpose\bf{x}$ where $\Phi$ are the eigenvectors from the eigendecomposition. Then the 
Formally, the whitening transformation of data $\text{\textbf x}$ with covariance $\Sigma$ is defined as $\text{\textbf w}{=}\Lambda^{-1/2}\Phi^\transpose\text{\textbf x}$, where $\Lambda$ and $\Phi$ are the eigenvalues and eigenvectors from the eigendecomposition of the covariance matrix $\Sigma$.
% ($\Sigma\Phi=\Phi\Lambda$)

Whitening impacts the model in two ways. The first is computational performance: since the structure $F$ is equal to $I$, it significantly saves computational costs in matrix multiplications~(eq.~\ref{eq:rpfg_vanilla}). The second is model quality, as both the computation of $\text{\textbf H}_{\text{opt}}$ and the compression phase, performed using Goldberger's K-means, benefit from working on spherized data.

\subsection{Optimal Bandwidth Stability}

Sometimes, observed data may produce zero-valued $\text{\textbf H}_{\text{opt}}$. Recall that we assume that the structure $F$ of $\text{\textbf H}_{\text{opt}}$ is well-approximated by the covariance of the entire model and that finding $\text{\textbf H}_{\text{opt}}$ amounts to finding $\beta_{opt}$ (section~\ref{sec:opt_band}). If this scaling factor is zero, e.g. caused by precision issues or by data distribution in the model, $\text{\textbf H}_{\text{opt}}$ will be zero. This means that no smoothing will be applied to the KDE. In this case, Dirac-deltas will remain with zero covariance, not allowing the use of the standard likelihood function. Since $\text{\textbf H}_{\text{opt}}$ is computed on whitened data, one solution for this problem is to detect these occurrences and set the bandwidth as the identity matrix, and then computing the backwards whitening transformation. 

%%%%%%%%%%%%%%%%%%%%%%%%%%%%%%%%%%%%%%%%%%%%%%%%%%%%%%%%%%%%%%%%%%%%%%%%%%%%%%%%
%%%%%%%%%%%%%%%%%%%%%%%%%%%%%%%%%%%%%%%%%%%%%%%%%%%%%%%%%%%%%%%%%%%%%%%%%%%%%%%%

\section{Using Diagonal Covariance Matrices}
\label{sec:diagcovmat}

Since a \gls{kde} is a linear combination of components, full covariances may not be necessary, even when features are not statistically independent: a second linear combination of (a higher number of) components with diagonal covariances is capable of modelling the correlations between features. 

With diagonal covariances, computational needs grow linearly with the number of dimensions, a very convenient aspect for high dimensionality applications. This is in contrast with quadratic growth for full covariances. Furthermore, for non-normalized data, diagonal covariances may improve stability, as certain relationships that could result in singular covariances are ignored. Our numerical stability strategies have a stronger impact with diagonal covariances: for data with flat or near-flat dimensions this is particularly relevant (section~\ref{sec:results}).

%%%%%%%%%%%%%%%%%%%%%%%%%%%%%%%%%%%%%%%%%%%%%%%%%%%%%%%%%%%%%%%%%%%%%%%%%%%%%%%%
%%%%%%%%%%%%%%%%%%%%%%%%%%%%%%%%%%%%%%%%%%%%%%%%%%%%%%%%%%%%%%%%%%%%%%%%%%%%%%%%

\section{Lazy Operation and Result Buffering}
\label{sec:lazy}

In theory, each time a sample is added, $\text{\textbf H}_{\text{opt}}$ should be updated. However, $\text{\textbf H}_{\text{opt}}$ is only needed for the compression phase or to evaluate the likelihood of a given sample. In other cases, computation of $\text{\textbf H}_{\text{opt}}$ may be postponed. This means that the cost of adding a sample is simply that of adding one Dirac-delta component to the mixture and updating the current weights. This can be done without impact on model quality. 

The determinant and the inverse of covariance matrices are needed to compute the likelihood function (which is called very often). Thus, it makes sense to save them, once they have been computed, and keep them until the covariance changes. Then, previous values are marked as invalid, but are not recomputed immediately. As with $\text{\textbf H}_{\text{opt}}$, it is advantageous to postpone these computations until they are actually needed, making the process of adding components to the mixture faster.

%%%%%%%%%%%%%%%%%%%%%%%%%%%%%%%%%%%%%%%%%%%%%%%%%%%%%%%%%%%%%%%%%%%%%%%%%%%%%%%%
%%%%%%%%%%%%%%%%%%%%%%%%%%%%%%%%%%%%%%%%%%%%%%%%%%%%%%%%%%%%%%%%%%%%%%%%%%%%%%%%

\section{Evaluation Setup}
\label{sec:eval_setup}

To evaluate our approach and to critically compare our results, we follow the same evaluation strategies of the original oKDE paper~\cite{kristan2011multivariate}. For intrinsic evaluation, we assess model quality by using the average negative log-likelihood and model complexity. For extrinsic evaluation, we assess the accuracy of a classification task. For the latter, we also compare our results with a discriminative approach, since it is typically better suited for classification tasks. We used the \gls{oisvm}~\cite{Orabona09,orabona2010line}. 

\begin{figure}[ht]
\centering
      \includegraphics[scale=0.45]{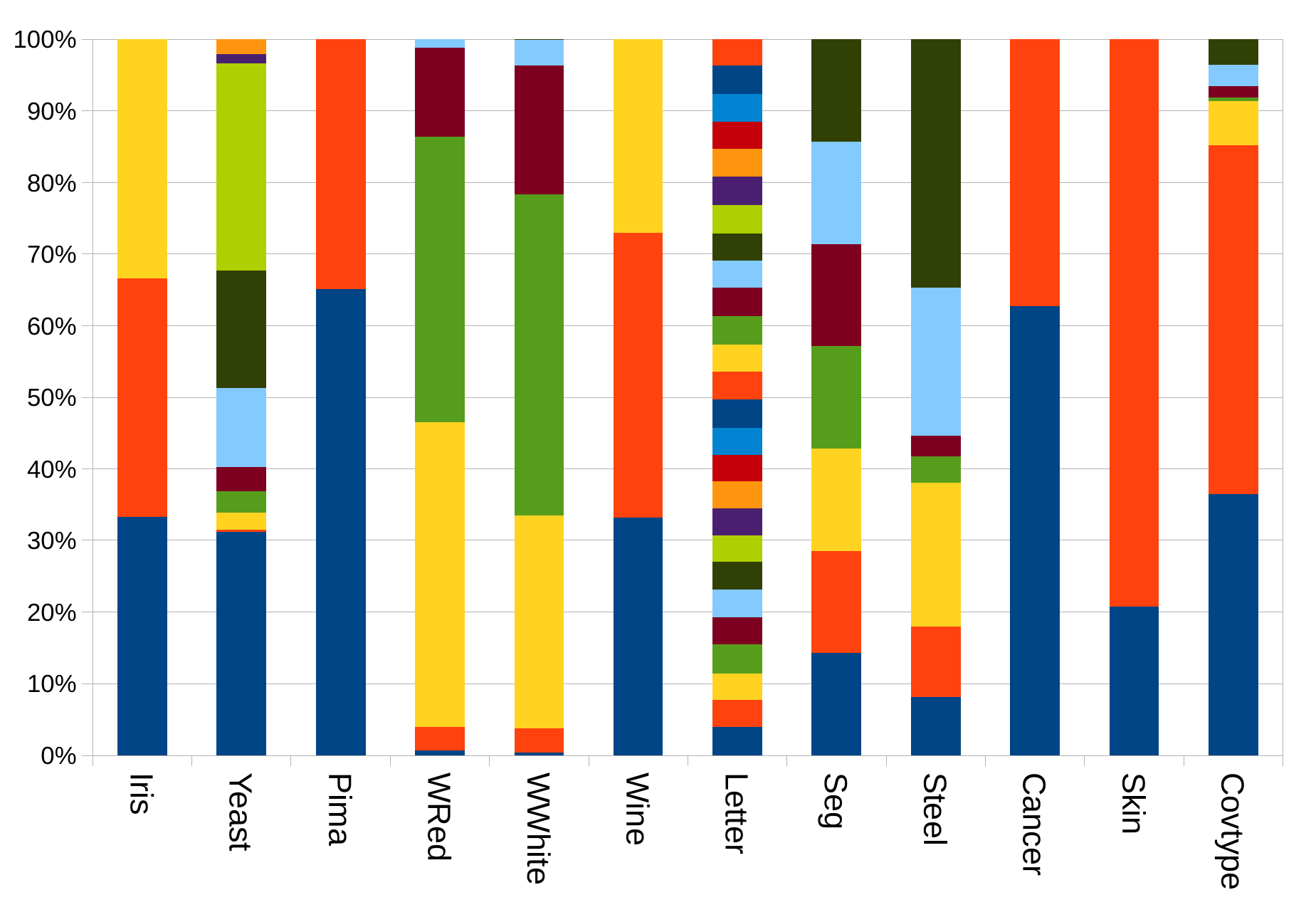}
  \caption{Class balancing accross the considered UCI datasets~\cite{Lichman:2013}.}
  \label{fig:uci_datasets_balancing}
\end{figure}

The datasets (Table~\ref{tab:uci_datasets} and Fig.~\ref{fig:uci_datasets_balancing}) and evaluation setup from the original oKDE paper were used. Since some of the datasets have been updated, we ran both approaches on the most recent versions. We also evaluated our approach on the L$^2$F Face Database, a high dimensional scenario (section~\ref{sec:faces}). For the online classification task, we randomly shuffled the data in each dataset and used 75\% of the data to train and the rest to test. For each dataset, to minimize impact of lucky/unlucky partitions, we generated 12 random shuffles. Since we wish to make minimal assumptions about data nature and distribution, we did not do any data preprocessing.

\begin{table}[ht]
  \begin{center}
\small\begin{tabular}{lrrr}
\hline
Dataset                  & N$_{\text S}$    & N$_{\text D}$ & N$_{\text C}$ \\ \hline
Iris                     & 150    & 4  & 3  \\
Yeast                    & 1484   & 8  & 10 \\
Pima                     & 768    & 8  & 2  \\
Red (wine quality, red)          & 1599   & 11 & 6  \\
White (wine quality, white)        & 4898   & 11 & 7  \\
Wine                     & 178    & 13 & 3  \\
Letter                   & 20000  & 16 & 26 \\
Seg (image segmentation) & 2310   & 19 & 7  \\
Steel                    & 1941   & 27 & 7  \\
Cancer (breast cancer)            & 569    & 30 & 2  \\
Skin                     & 245057 & 3  & 2  \\
Covtype                  & 581012 & 10 & 7  \\ \hline \\
\end{tabular}\normalsize
\caption[Properties of the UCI datasets]{Properties of the considered UCI datasets~\cite{Lichman:2013}, denoted by number of samples (N$_{\text S}$), number of dimensions  (N$_{\text D}$), and number of classes (N$_{\text C}$)}
    \label{tab:uci_datasets}
  \end{center}
\end{table}

For the \gls{kde} approaches, each class is represented by the model built from its training samples: for oKDE, each model is initialized with 3 samples before adding the rest, one at a time; in our approach, due to lazy operation, there is no need for initialization, and each model is trained by adding one sample at a time from the very start. To evaluate classification performance we used $\hat{y}{=}\argmax_k p(\text{\bf x} | c_k)p(c_k)$.

For \gls{oisvm}, we trained $k$ binary classifiers, using training samples from a class as positive examples, and the other training samples as negative examples. Then, we followed a 1-vs-All approach, $\hat{y}{=}\argmax_k f_k(\text{\bf x})$, in which $f_k(\text{\bf x})$ is the confidence score of the $k$-th binary classifier for $\text{\bf x}$. A second order polynomial kernel was used with \textsl{gamma} and \textsl{coef0} parameters set to 1. The complexity parameter \textsl{C} was also set to 1.

\subsection{L$^2$F Face Database}
\label{sec:faces}

The L$^2$F Face Database consists of 30,000 indoor free pose face images from 10 subjects (3,000 images per subject). Capture was performed using a PlayStation Eye camera (640x480 pixel). A Haar cascade~\cite{viola2001rapid,opencv_library} detected frontal face poses, but also some high inclination and head rotation poses. The square face regions were cropped and resized to 64x64 pixels. During capture, each subject was asked to behave normally and avoid being static, to avoid restrictions on facial pose or expression. Since faces tend to be centered in the cropped square, a fixed mask, roughly selecting parts of eyes, nose, and cheeks, was applied to the crops, producing 128-pixel vectors.

%%%%%%%%%%%%%%%%%%%%%%%%%%%%%%%%%%%%%%%%%%%%%%%%%%%%%%%%%%%%%%%%%%%%%%%%%%%%%%%%
%%%%%%%%%%%%%%%%%%%%%%%%%%%%%%%%%%%%%%%%%%%%%%%%%%%%%%%%%%%%%%%%%%%%%%%%%%%%%%%%

\section{Results}
\label{sec:results}

Tests were run on an Intel Xeon E5530@2.40GHz, 48GB RAM, Linux openSUSE 13.1. oKDE and \gls{oisvm} were run on MATLAB R2013a with {\em -nosplash -nodesktop -nodisplay \mbox{-nojvm}}. Our approach (xokde++) was compiled enabling all vectorizing options available on our CPU architecture. In the tables, ``{---}'' indicates that the model could not be built.

\subsection{Model Quality}
As a proxy for estimation quality, we use the average negative log-likelihood and the average complexity of the models. Tables~\ref{tab:cplx_nll_loglike} and~\ref{tab:cplx_nll_complexity} present results after observing all samples. 

\begin{table}[ht]
  \begin{center}
\small\begin{tabular}{l|rr|rr|rr}
   & \multicolumn{2}{|c|}{xokde++} & \multicolumn{2}{|c|}{xokde++/d} & \multicolumn{2}{|c}{oKDE} \\\hline
Dataset   & -$\mathcal{L}$ & $\sigma$ & -$\mathcal{L}$ & $\sigma$ & -$\mathcal{L}$ & $\sigma$   \\    \hline 
Iris         & 7.4   & 1.7       & 3.3       & 1.3  & 7.8   & 1.6  \\
Yeast        & -13.2 & 0.5       & -14.3     & 0.3  & -7.1 & 1.7  \\
Pima         & 29.3  & 0.5       & 27.6      & 0.2  & 31.0  & 0.7  \\
Red          & -0.2  & 0.6       & 0.2       & 0.3  & 14.7   & 4.2  \\
White        & 0.5   & 0.2       & 1.6       & 0.2  & 55.2   & 54.3  \\
Wine         & 51.1  & 4.3       & 33.0      & 3.4  & 64.9  & 18.5 \\
Letter       & 17.9  & 0.1       & 19.9      & 0.1  & 18.2  & 0.1  \\
Seg          & 25.7  & 0.8       & 31.0      & 1.0  & 156.6 & 33.7 \\
Steel        & ---   & ---       & 85.2      & 4.3  & --- & --- \\
Cancer       & -25.6 & 5.7       & -16.4     & 2.4  & 433.9 & 33.7 \\
Skin         & 13.5  & 0.3       & 13.9      & 0.2  & 13.4  & 0.2  \\
Covtype      & 51.9  & 0.1       & 55.9      & 0.0  & 52.6  & 0.1  \\ \hline
\end{tabular}    \normalsize
\caption[Average negative log-likelihood]{Average negative log-likelihood -$\mathcal{L}$ ($\sigma$ = one standard deviation).}
    \label{tab:cplx_nll_loglike}
  \end{center}
\end{table}

\begin{table}[ht]
  \begin{center}
\small\begin{tabular}{l|rr|rr|rr}
   & \multicolumn{2}{|c|}{xokde++} & \multicolumn{2}{|c|}{xokde++/d} & \multicolumn{2}{|c}{oKDE} \\\hline
Dataset   & $K$ & $\sigma$ & $K$ & $\sigma$ & $K$ & $\sigma$   \\    \hline 
Iris     & 28         & 3    & 22        & 3    & 28    & 3   \\
Yeast    & 31         & 15   & 30        & 19   & 31    & 15  \\
Pima     & 62         & 9    & 42        & 19   & 64    & 7   \\
Red      & 39         & 23   & 53        & 37   & 40    & 24  \\
White    & 31         & 24   & 54        & 40   & 36    & 25  \\
Wine     & 44         & 7    & 44        & 7    & 45    & 7   \\
Letter   & 65         & 12   & 42        & 9    & 66    & 14  \\
Seg      & 49         & 12   & 51        & 24   & 52    & 11  \\
Steel    & ---        & ---  & 20        & 7    & ---   & --- \\
Cancer   & 40         & 7    & 153       & 11   & 50    & 12  \\
Skin     & 8          & 2    & 10        & 2    & 8     & 2   \\
Covtype  & 18         & 5    & 17        & 5    & 20    & 5   \\ \hline
\end{tabular}    \normalsize
\caption[Average model complexity]{Average number of components ($K$) ($\sigma$ = one standard deviation).}
    \label{tab:cplx_nll_complexity}
  \end{center}
\end{table}

Relative to \gls{okde}, xokde++ produces models with similar complexity but with lower average negative log-likelihood (better fits).
This is clear in datasets where some dimensions have very little variance while others have very large variance. The ``steel'', ``segmentation'', and ``cancer'' datasets are examples of this kind of problem. The numerical stability methods in xokde++ allow recovery from these situations (section~\ref{sec:degen_cov_mat}).

\subsection{Classifier Accuracy}
\label{sec:accuracy}

Table~\ref{tab:clf_acc} shows that xokde++ achieves better performance in 7 out of 12 datasets, and lower performance in only 3 datasets. Some of the results for oKDE are different than those reported in the original paper, perhaps due to the fact that no dataset was balanced or normalized for unit variance. We intentionally used the datasets with no further processing to study a more realistic online operation scenario, where the sample distribution is not known. These results demonstrate the numerical robustness of xokde++ in handling non-normalized datasets. \gls{oisvm} typically outperforms the generative approaches, but only slightly. However, it shares some of the oKDE numerical instabilities. 

\begin{table}[ht]
  \begin{center}
\small\begin{tabular}{l|rr|rr|rr|rr}
   & \multicolumn{2}{|c|}{xokde++} & \multicolumn{2}{|c|}{xokde++/d} & \multicolumn{2}{|c}{oKDE} & \multicolumn{2}{|c}{OISVM} \\\hline
Dataset  & $A$ & $\sigma$ & $A$ & $\sigma$ & $A$ & $\sigma$ & $A$ & $\sigma$   \\    \hline 
Iris     & 96.4    & 2.7  & 95.0    & 3.4 & 96.4 & 2.4  & 97.1  & 2.1  \\
Yeast    & 49.7    & 2.3  & 48.1    & 1.6 & 50.6 & 3.3  & 59.2  & 2.2  \\
Pima     & 67.8    & 3.4  & 70.1    & 3.9 & 69.7 & 2.9  & 76.9  & 2.5  \\
Red      & 62.0    & 2.5  & 54.6    & 1.9 & 56.9 & 6.3  & 58.3  & 1.9 \\
White    & 49.9    & 1.3  & 42.4    & 1.7 & 44.9 & 10.6 & 53.2  & 1.6 \\
Wine     & 97.7    & 1.4  & 98.5    & 1.8 & 93.9 & 6.1  & 96.8  & 2.8  \\
Letter   & 95.8    & 0.2  & 93.4    & 0.4 & 95.8 & 0.2  & 93.0  & 0.4  \\
Seg      & 91.5    & 1.1  & 89.4    & 1.2 & 75.0 & 5.3  & 95.0  & 0.9 \\
Steel    & ---     & ---  & 56.9    & 9.0 & ---  & ---  & ---   & ---  \\
Cancer   & 94.8    & 1.7  & 96.2    & 1.7 & 52.8 & 12.0 & 95.9  & 0.8  \\
Skin     & 99.6    & 0.1  & 99.4    & 0.0 & 99.7 & 0.1  & 99.8  & 0.0  \\
Covtype  & 52.0    & 1.2  & 51.6    & 0.6 & 68.0 & 0.9  & {---} & {---}    \\ \hline
\end{tabular}\normalsize
    \caption[Classification Accuracy]{Average classification accuracy ($A$) ($\sigma$ = one standard deviation).}
    \label{tab:clf_acc}
      \end{center}
\end{table}

The high dimensionality scenario was tested using the L$^2$F Face Database. \gls{okde}, \gls{oisvm}, and xokde++ (with full covariances), were unable to build models for this dataset. xokde++ (with diagonal covariances) obtained an accuracy of 94.1\%, very close to the accuracy of 95.4\% obtained by a batch SVM.

\subsection{Time and Memory Performance}

xokde++ produces models with similar or better quality than \gls{okde}. Moreover, it does so at lower computational cost. Tables~\ref{tab:mem_perf} and~\ref{tab:time_perf} show the time and memory needed to train and test each dataset: xokde++ achieves speedups from 3 to 10; with diagonal covariances, speedups range from 11 to 40. Regarding memory, xokde++ uses at most 10\% of the memory required by \gls{okde}. This difference is more critical in large datasets such as the ``skin'' and ``covtype''. For these datasets, oKDE needed 913MB and 5064MB, while xokde++ needed only 83MB and 361MB, respectively. 

Regarding memory, \gls{oisvm} performs poorly when compared with the other approaches. For the ``letter'' dataset, with 26 classes, the memory needed by \gls{oisvm} was 4 times the required by \gls{okde} and 43 and 78 times more than xokde++ with full and diagonal covariances, respectively. For large datasets, time performance also degrades: the ``skin'' dataset took nearly twice the time it took for \gls{okde}. This is even clearer for the ``covtype'' dataset, which has 500,000 samples: 7 days of computation were not enough to complete training a single shuffle, showing that it was unfeasible to complete the 12 shuffles of our evaluation setup in an acceptable time frame. 

\begin{table}[ht]
  \begin{center}
\small\begin{tabular}{l|rr|rr|r|rr}
   & \multicolumn{2}{|c|}{xokde++} & \multicolumn{2}{|c|}{xokde++/d} & \multicolumn{1}{|c}{oKDE} & \multicolumn{2}{|c}{OISVM} \\\hline
Dataset  & MB & $\rho$\% & MB & $\rho$\% & MB  & MB & $\rho$\%   \\    \hline 
Iris        & 4.6   & 3.8 & 4.5   & 3.7 & 123.0  & 116.6 & 94.8  \\
Yeast       & 6.8   & 3.9 & 5.8   & 3.4 & 173.1  & 148.9 & 86.0  \\
Pima        & 6.2   & 3.9 & 5.2   & 3.3 & 157.5  & 124.8 & 79.3  \\
Red         & 8.1   & 5.6 & 6.2   & 4.2 & 145.4  & 162.2 & 111.5 \\
White       & 10.3  & 5.1 & 8.3   & 4.1 & 204.3  & 228.4 & 111.8 \\
Wine        & 6.3   & 4.4 & 4.8   & 3.3 & 144.3  & 120.2 & 83.3  \\
Letter      & 42.6  & 9.8 & 23.2  & 5.4 & 433.3  & 1823.9 & 420.9 \\
Seg         & 15.7  & 7.7 & 7.7   & 3.8 & 203.9  & 192.3 & 94.3  \\
Steel       & ---   & --- & 7.2   & 4.0 & ---    & ---  & --- \\
Cancer      & 10.6  & 5.5 & 6.6   & 3.4 & 193.1  & 141.8 & 73.4  \\
Skin        & 83.1  & 9.1 & 83.2  & 9.1 & 913.3  & 2255.1 & 246.9 \\
Covtype     & 361.2 & 7.1 & 360.4 & 7.1 & 5064.8 & {---} & {---}  \\ \hline
  \end{tabular}\normalsize
    \caption[Memory Performance]{Results for memory performance (MB). $\rho$ is the relative usage against the oKDE baseline.}
    \label{tab:mem_perf}
  \end{center}
\end{table}

\begin{table*}[ht]
  \begin{center}
\small\begin{tabular}{l|rrr|rrr|rr|rrr}
   & \multicolumn{3}{|c|}{xokde++} & \multicolumn{3}{|c|}{xokde++/d} & \multicolumn{2}{|c|}{oKDE} & \multicolumn{3}{|c}{OISVM} \\\hline
Dataset   & time & $\sigma$ & $\rho$ & time & $\sigma$ & $\rho$  & time & $\sigma$  & time & $\sigma$ & $\rho$  \\    \hline 
Iris       & 0.5    & 0.1    & 11.0 & 0.1       & 0.0   & 34.2 & 5.0     & 0.7    & 0.2     & 0.1    & 26.7   \\
Yeast      & 39.4   & 3.2    & 4.5  & 6.5       & 0.7   & 27.1 & 177.2   & 8.1    & 7.7     & 0.3    & 23.1   \\
Pima       & 16.9   & 2.1    & 7.9  & 3.3       & 0.3   & 40.1 & 133.9   & 2.3    & 1.5     & 0.4    & 90.1   \\
Red        & 61.0   & 7.7    & 5.4  & 14.4      & 0.7   & 22.6 & 326.4   & 27.5   & 10.6    & 1.8    & 30.7   \\
White      & 130.7  & 4.0    & 5.8  & 39.0      & 1.7   & 19.6 & 764.1   & 51.3   & 37.2    & 3.6    & 20.5   \\
Wine       & 3.0    & 0.5    & 3.2  & 0.5       & 0.1   & 21.4 & 9.7     & 0.6    & 0.3     & 0.1    & 29.4   \\
Letter     & 2119.0 & 66.7   & 3.0  & 191.7     & 3.5   & 33.7 & 6452.9  & 191.9  & 5225.4  & 1341.0 & 1.2    \\
Seg        & 296.8  & 26.1   & 1.7  & 45.0      & 6.1   & 11.5 & 515.7   & 21.5   & 84.1    & 12.0   & 6.1    \\
Steel      & ---    & ---    & ---  & 8.7       & 0.5   & ---  & ---     & ---    & ---     & ---    & ---    \\
Cancer     & 52.6   & 6.9    & 2.6  & 11.7      & 1.1   & 11.6 & 135.6   & 11.2   & 0.6     & 0.1    & 221.8  \\
Skin       & 572.9  & 204.5  & 11.5 & 192.2     & 40.9  & 34.2 & 6565.1  & 1471.0 & 10522.5 & 2144.0 & 0.6    \\
Covtype    & 9803.8 & 1614.9 & 2.6  & 1156.6    & 124.5 & 22.2 & 25713.3 & 2081.3 & {---}   & {---}  & {---}  \\ \hline
\end{tabular} \normalsize  
\caption[Time Performance]{Results for time performance (seconds) ($\sigma$ = one standard deviation). $\rho$ is the speedup over the oKDE baseline.}
\label{tab:time_perf} 
  \end{center}
\end{table*}

\subsection{Performance of Full vs. Diagonal Covariances}

Table~\ref{tab:comparison_full_diag} presents a detailed comparison of the two xokde++ variants, regarding model quality and computational needs. Using diagonal covariances produces models with slightly higher complexity at a small loss in model quality, while needing less memory and greatly improving computational time performance. For the ``steel'' dataset, diagonal covariances also allowed the model to cope with feature dimensions with low variance, thus avoiding the severe numerical instability found when using full covariances, where the lack of feature variance frequently produced singular matrices. 

For full covariances, the number of variables increases quadratically with the number of dimensions and datasets with higher dimensionality will benefit from using diagonal covariances. Even though the diagonal approach tends to need more components, partially offsetting memory savings, the average memory usage of the diagonal approach is just 78.5\% of the needed for full covariances. We observed an improvement of this number when we considered more than 15 dimensions: the average memory usage falls to 73\% with only small losses in accuracy, around 0.72\% (absolute). Considering all datasets, we obtain an average value of 1.67\% (absolute) accuracy loss against full covariance models.

\begin{table*}[ht]
  \begin{center}
\small\begin{tabular}{l|rrrrr|rrrrr|rrrr}
        & \multicolumn{5}{|c|}{xokde++}         & \multicolumn{5}{c|}{xokde++/d}     &    \multicolumn{4}{|c}{comparison}    \\\hline
Dataset & $A_f$\%  & $K_f$ & -$\mathcal{L}_f$ & $T_f$ & $M_f$ &  $A_d$\%  & $K_d$ & -$\mathcal{L}_d$ & $T_d$ & $M_d$  & $\Delta A$\%  & $\Delta K$ & $S$ & $M$\% \\\hline
Iris    & 96.4 & 28  & 7.4   & 0.5    & 4.6   & 95.0 & 22  & 3.3   & 0.1    & 4.5   & -1.4 & -7  & 3.1  & 97.9  \\
Yeast   & 49.7 & 31  & -13.2 & 39.4   & 6.8   & 48.1 & 30  & -14.3 & 6.5    & 5.8   & -1.6 & -1  & 6.0  & 85.2  \\
Pima    & 67.8 & 62  & 29.3  & 16.9   & 6.2   & 70.1 & 42  & 27.6  & 3.3    & 5.2   & 2.3  & -21 & 5.1  & 83.5  \\
Red     & 62.0 & 39  & -0.2  & 61.0   & 8.1   & 54.6 & 53  & 0.2   & 14.4   & 6.2   & -7.4 & 14  & 4.2  & 75.8  \\
White   & 49.9 & 31  & 0.5   & 130.7  & 10.3  & 42.4 & 54  & 1.6   & 39.0   & 8.3   & -7.5 & 22  & 3.4  & 80.8  \\
Wine    & 97.7 & 44  & 51.1  & 3.0    & 6.3   & 98.5 & 44  & 33.0  & 0.5    & 4.8   & 0.8  & 0   & 6.6  & 75.8  \\
Letter  & 95.8 & 65  & 17.9  & 2119.0 & 42.6  & 93.4 & 42  & 19.9  & 191.7  & 23.2  & -2.4 & -23 & 11.1 & 54.5  \\
Seg     & 91.5 & 49  & 25.7  & 296.8  & 15.7  & 89.4 & 51  & 31.0  & 45.0   & 7.7   & -2.2 & 2   & 6.6  & 48.7  \\
Steel   & ---  & --- & ---   & ---    & ---   & 56.9 & 20  & 85.2  & 8.7    & 7.2   & ---  & --- & ---  & ---   \\
Cancer  & 94.8 & 40  & -25.6 & 52.6   & 10.6  & 96.2 & 153 & -16.4 & 11.7   & 6.6   & 1.5  & 113 & 4.5  & 61.8  \\
Skin    & 99.6 & 8   & 13.5  & 572.9  & 83.1  & 99.4 & 10  & 13.9  & 192.2  & 83.2  & -0.1 & 3   & 3.0  & 100.1 \\
Covtype & 52.0 & 18  & 51.9  & 9803.8 & 361.2 & 51.6 & 17  & 55.9  & 1156.6 & 360.4 & -0.4 & -2  & 8.5  & 99.8  \\ \hline
\end{tabular}\normalsize
    \caption[Full vs Diagonal covariance performance]{Comparison of using full vs. diagonal covariances, on accuracy ($A$\%), number of model components ($K$), negative log-likelihood (-$\mathcal{L}$), time ($T$ in seconds), and memory ($M$ in megabytes). In the comparison columns, $\Delta \bullet=\bullet_d-\bullet_f$, $S=T_f/T_d$ (speed up), and $M=M_d/M_f$ (relative memory use).}
        \label{tab:comparison_full_diag}
  \end{center}
\end{table*}

%%%%%%%%%%%%%%%%%%%%%%%%%%%%%%%%%%%%%%%%%%%%%%%%%%%%%%%%%%%%%%%%%%%%%%%%%%%%%%%%
%%%%%%%%%%%%%%%%%%%%%%%%%%%%%%%%%%%%%%%%%%%%%%%%%%%%%%%%%%%%%%%%%%%%%%%%%%%%%%%%

\section{Conclusions}
\label{sec:conclusions}

xokde++ is a state-of-the-art online \gls{kde} approach. It is efficient, numerically robust, and able to handle high dimensionality. Model quality comparable to that of oKDE, while needing significantly less memory and computation time. The numerical stability improvements allow xokde++ to cope with non-normalized data and achieve an average classifier accuracy of 78\% (against oKDE's 73\%). This is a 6.5\% average improvement over the baseline. Furthermore, it continues to achieve good results where other approaches fail to even convege~(section~\ref{sec:accuracy}). From a software standpoint, xokde++ is extensible, due to its pure template library nature, as shown by the use of diagonal covariances, which were easily implemented and provide aditional time and memory efficiency while sacrificing little model quality.

%%%%%%%%%%%%%%%%%%%%%%%%%%%%%%%%%%%%%%%%%%%%%%%%%%%%%%%%%%%%%%%%%%%%%%%%%%%%%%%%
%%%%%%%%%%%%%%%%%%%%%%%%%%%%%%%%%%%%%%%%%%%%%%%%%%%%%%%%%%%%%%%%%%%%%%%%%%%%%%%%

Besides speed and memory efficiency, one central contribution of our approach is its adaptability and extensibility. Since it was implemented in an OOP language (C++), the templated classes allow for easy extension, as shown by switching between full and diagonal covariances. Other parts of the code are also easily extended, e.g., using triangular covariances, or replacing the Gaussian \gls{pdf} with a different one. Similarly, changing the compression phase is also a simple process.

Future work to improve computational performance includes high-performance computing adaptations. Other improvements include changing the hierarchical compression approach, to another, more paralelizable approach such as pairwise merging of components~\cite{runnalls2007kullback}.
Moreover, model quality can also be improved. Instead of using full or diagonal covariances, which are either too large for high dimensionality or too restrictive for certain non-linearly independent data, covariances could be approximated with low-rank perturbations~\cite{magdon2009approximating}, or using Toeplitz matrices~\cite{pasupathy1992gaussian,cai2014estimating}. Finally, improvements on numerical stability and model quality may be obtained by avoiding correction of degenerate covariances and, instead, compute pseudo-inverses and pseudo-log-determinants~\cite{mikheev2006multidimensional}.



%\bibliographystyle{IEEEtran}

%%%%%%%%%%%%%%%%%%%%%%%%%%%%%%%%%%%%%%%%%%%%%%%%%%%%%%%%%%%%%%%%%%%%%%%%%%%%%%%%
%%%%%%%%%%%%%%%%%%%%%%%%%%%%%%%%%%%%%%%%%%%%%%%%%%%%%%%%%%%%%%%%%%%%%%%%%%%%%%%%

\section{Acknowledgements}

This work was supported by national funds through Fundação para a Ciência e a Tecnologia (FCT) with reference \linebreak UID/CEC/50021/2013.

\bibliographystyle{model2-names}
\bibliography{document}

\end{document}